\RequirePackage{fix-cm}
\documentclass{svjour3}                     % onecolumn (standard format)
\smartqed  % flush right qed marks, e.g. at end of proof
\usepackage{graphicx}
\usepackage{cite}
\usepackage{amsmath,amssymb,amsfonts}
\usepackage{algorithmic}
\usepackage{caption}
\usepackage{textcomp}
\usepackage{hyperref}
\usepackage{float}
\usepackage[utf8]{inputenc}
\journalname{Nonlinear Dynamics}
\begin{document}

%------------------------------------------------------------------
\title{Empirical Mode Modeling}
\subtitle{A data-driven approach to recover and forecast nonlinear dynamics from noisy data}

\titlerunning{Empirical Dynamic Modeling} % if too long for running head

\author{ Joseph Park \and Gerald M Pao \and
         George Sugihara \and Erik Stabenau \and Thomas Lorimer }


\institute{Joseph Park \and Erik Stabenau \at
           U.S. Department of the Interior \\
           South Florida Natural Resources Center, Homestead FL 33031 USA\\
           \email{JosephPark@IEEE.org}
           \and
           Gerald M Pao \at
           Salk Institute for Biological Studies\\
           MCBL-4, La Jolla CA 92037 USA\\
           \email{pao@Salk.edu}
           \and
           George Sugihara \and Thomas Lorimer \at
           Scripps Institution of Oceanography\\
           University of California San Diego, La Jolla CA 92037 USA\\
           }

\date{Received: date}

\maketitle

%------------------------------------------------------------------
\begin{abstract}
%------------------------------------------------------------------
Data-driven, model-free analytics are natural choices for discovery and forecasting of complex, nonlinear systems. Methods that operate in the system state-space require either an explicit multidimensional state-space, or, one approximated from available observations.  Since observational data are frequently sampled with noise, it is possible that noise can corrupt the state-space representation degrading analytical performance.  Here, we evaluate the synthesis of empirical mode decomposition with empirical dynamic modeling, which we term empirical mode modeling, to increase the information content of state-space representations in the presence of noise.  Evaluation of a mathematical, and, an ecologically important geophysical application across three different state-space representations suggests that empirical mode modeling may be a useful technique for data-driven, model-free, state-space analysis in the presence of noise.
\end{abstract}

%------------------------------------------------------------------
\section{Introduction}
\label{sec:introduction}
%------------------------------------------------------------------
Evolution of science and technology is episodically redirected as our understanding improves.  For example, the late 19$\mathrm{^{th}}$ to mid-20$\mathrm{^{th}}$ Century recognition of nonlinear dynamical systems as not purely stochastic, or, purely deterministic was both troubling and opportune.  The subsequent acknowledgment of emergent behaviors as a {\em property} of such systems can be considered such a redirection\cite{Anderson1972}.  Progress along this front established dynamical systems theory as a new branch of science \cite{Journals2020}. Accordingly, it is now well-accepted that canonical statistical and parametric equation-based models applied to nonlinear systems, which turn out to be nearly ubiquitous in complex technologies and nature, are often problematic \cite{DeAngelis2015}.

Currently, machine learning is advancing new directions in data and system analysis. Techniques such as manifold learning \cite{Lin2015} and diffusion maps \cite{Coifman2006} rely on a {\em data-driven}, inductive approach, wherein the data-itself reveals underlying dynamics and behavioral complexities, rather than a presumptive, model-based deductive analysis. As effective as these methods are at finding patterns and forecasting, their scientific utility is less clear as the patterns uncovered do not typically translate into mechanistic insight. For example, identifying intervariable interactions and causal relationships between system variables. 

A possible exception is a set of dynamically-based, data-driven analytics termed empirical dynamic modeling (EDM), which have developed into a useful toolset for system analysis and forecasting\cite{Sugihara1990, Sugihara1994, Dixon1999, Sugihara2012, Ye2016}. EDM operates in multidimensional state-space, either from an explicit multidimensional data set, or, from a diffeomorphic reconstruction of the state-space, for example, by application of Takens embedding theorem \cite{Takens1981, Deyle2011}.  EDM is not based on parametric presumptions, fitting statistics, or, specifying equations; representing a data-driven approach amenable to nonlinear dynamics as noted by DeAngelis \cite{DeAngelis2015}.  The unfamiliar reader is referred to the lucid EDM introduction in reference \cite{Chang2017}.

Since these methods are data-driven, it is possible that observational noise can interfere with accurate state-space representations, degrading inference and forecast skill.  We therefore seek methods to improve state-space representations derived from noisy observations, and here, turn to empirical mode decomposition (EMD). 

EMD decomposes oscillatory signals into scale-dependent modes termed intrinsic mode functions (IMF) without constraints of linearity or stationarity as presumed by Fourier, wavelet or Eigen decomposition.  Application of the Hilbert transform to IMFs provides time-dependent instantaneous frequency estimates, with the combination of EMD and IMF Hilbert spectra constituting the Hilbert-Huang transform (HHT) \cite{Huang2008}.  IMFs are particularly astute at isolating physically-meaningful dynamics \cite{Huang2008}, and, their data-driven isolation of noise is a cornerstone of EMD utility \cite{Dai2019}. An introduction and review are found in references \cite{Huang2008} and \cite{Looney2015}.

Since dynamics of interest are often low-dimensional and contained within a bounded state-space, univariate projections of state-space trajectories are often expressed as oscillatory signals.  EMD provides a model-free and natural way to filter noise and isolate physically relevant modes of oscillatory time series, while EDM enables data-driven state-space metrics for forecasting and quantifying cross-variable interactions. 

Here, we combine the model-free modal decomposition of EMD with the multivariate state-space representation inherent in EDM to improve forecasting and discovery in the presence of noisy or confounded observations. We term this empirical mode modeling (EMM). 

We demonstrate the utility of this synthesis in two distinct applications.  First a nonlinear, chaotic mathematical system, the Rössler attractor, and second, a nonlinear geophysical system with ecological importance, salinities in Florida Bay within Everglades National Park.

%------------------------------------------------------------------
\subsection{Model fitness and application}
\label{sec:modelfitness}
%------------------------------------------------------------------
Following Granger \cite{Granger1969}, we adopt a position that model predictability reflects the amount of information a model contains regarding a dynamical system. We therefore use model fidelity expressed as the Pearson correlation $\rho$ between observations and model predictions as a fitness metric. Note that the models themselves can be equation-free and nonlinear, and, that other metrics such as mutual information are equally valid. 

When models are used in an operational forecast system it is typical to assess model fitness on {\it out-of-sample} data, that is, the model is trained on a subset of available data with fitness assessed on a prediction (validation) set disjoint from the training data.  Out-of-sample prediction characterizes model stability and the ability to generalize to untrained data.

{\it In-sample} models, where the training and prediction sets overlap, are informative when discovering/assessing the ability of the model to represent system dynamics. For example, as a function of state-space variables and their interactions, or, in response to noise/confounded states.  We use both formats to assess model fitness and dynamical information content.

%---------------------------------------------------------------------------
\section{Empirical Mode Modeling}
%---------------------------------------------------------------------------
\label{sec:methods}
The basis of EMM is to create a multidimensional state-space from empirical mode decomposition of observational time series.  The resultant IMFs, or subsets of IMFs representing the system state-space, are then analyzed within the empirical dynamic modeling framework.  The reader is referred to reference \cite{Looney2015} for details of empirical mode decomposition, and reference \cite{Chang2017} for empirical dynamic modeling. 

Generically, EMM can be implemented with the following steps:

\begin{enumerate}
\item Determine the signal-to-noise ratio (SNR) of the time series.  If the SNR is less than 3 dB, EMM may provide improvements in state-space representation and forecasting. 
  
\item Apply empirical mode decomposition to the time series creating a set of intrinsic mode functions (IMFs).  This can be done for multiple time series. 

\item Optionally select IMFs that best represent the dynamics. This is an ongoing area of research.  Here, we use two methods:
  \begin{enumerate}
  \item Multiview embedding as described in reference \cite{Ye2016}, selecting the set of IMFs that maximize model predictability as exemplified in section \ref{IMFSelection}.
  \item Select IMFs based on spectral filtering, or, manual inspection to remove high frequency noise or low frequency oscillations external to the dynamics.  This is shown in section \ref{stateSpaceModels}. 
  \end{enumerate}

\item Apply empirical dynamic modeling using the selected IMFs as a multivariable state-space.  Application of EDM to multivariable state-space is described in reference \cite{Dixon1999}.  EDM analysis can be performed using simplex\cite{Sugihara1990}, sequential locally weighted global linear maps (S-maps)\cite{Sugihara1994}, or convergent cross mapping\cite{Sugihara2012}. 
  
\end{enumerate}

%---------------------------------------------------------------------------
\section{Rössler Dynamics}
%---------------------------------------------------------------------------
The Rössler attractor is a 3-dimensional coupled dynamical system known to exhibit chaotic dynamics\cite{Rossler1976}.  Our implementation is defined as:
\begin{align}
 \label{equ:Rosslerx}
 \frac{dx}{dt} = -y-z\\
 \label{equ:Rosslery}
 \frac{dy}{dt} = x + ay\\
 \label{equ:Rosslerz}
 \frac{dz}{dt} = b + z(x-c)
\end{align}
\noindent with constants $a$ = 0.4, $b$ = 0.4, $c$ = 4, initial state values $x_0$ = 1, $y_0$ = 0, $z_0$ = 1, time increment = 0.01, and time span $T$ = [0-500].  We then subsample the integrated solutions of equations \ref{equ:Rosslerx} - \ref{equ:Rosslerz} at a time increment of 0.1, and ignore the time span from 0 to 200 resulting in a 3-D time series of 3000 points from time 200 to 500.  Multispectral noise ($N$) is generated following reference \cite{Timmer1995} as a combination of brown and pink noise:
\begin{equation}
  N = A~( B~ p_n + C~ b_n )
  \label{equ:Noise}
\end{equation}
\noindent where $p_n$ is "pink" noise with power spectral density $1/f$, $b_n$ is low frequency "Brownian" noise with power spectrum $1/f^2$, where $f$ is frequency, and, noise amplitude $A$ ranges from 0 to 64; $B$ = 0.5, $C$ = 1. Corresponding signal-to-noise ratios are listed in Table \ref{table:SNR}.

\begin{table}[h!]
\centering
\caption{Rössler signal-to-noise ratios}
\label{table:SNR}
\begin{tabular}{r|cccccccccc}
\hline\noalign{\smallskip}
Noise A  & 1 & 2 & 4 & 8 & 12 & 16 & 24 & 32 & 48 & 64\\
\noalign{\smallskip}\hline\noalign{\smallskip}
SNR dB & 10.08&7.00&4.12&1.07&-0.73&-1.89&-3.71&-4.94&-6.69&-7.95\\
\noalign{\smallskip}\hline
\end{tabular}
\end{table}

\begin{figure}
\centering
\includegraphics[width=1\textwidth]{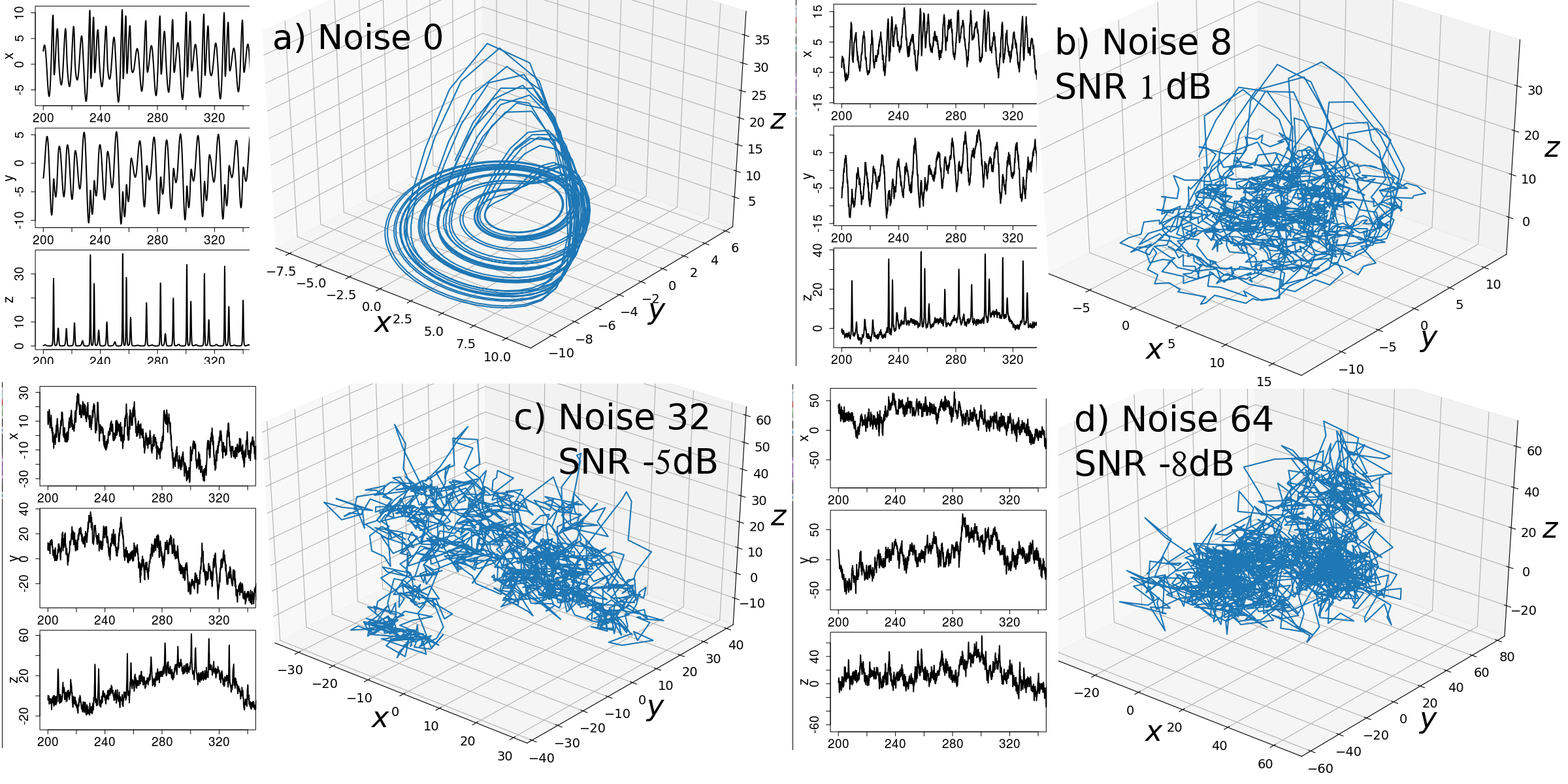}
\caption{State-space time series and trajectories of Rössler dynamics with multispectral noise.}
\label{fig:RosslerNoise}
\end{figure}

Fig. \ref{fig:RosslerNoise} shows four examples of the Rössler state-space with different levels of additive noise.  It is clear that as noise increases, trajectories in the state-space become corrupted and entangled, and, we expect that methods such as manifold learning and EDM that rely on state-space representations of system dynamics can perform poorly as noise increases.

%---------------------------------------------------------
\subsection{Empirical mode decomposition}
\begin{figure}
\centering
\includegraphics[width=0.8\textwidth]{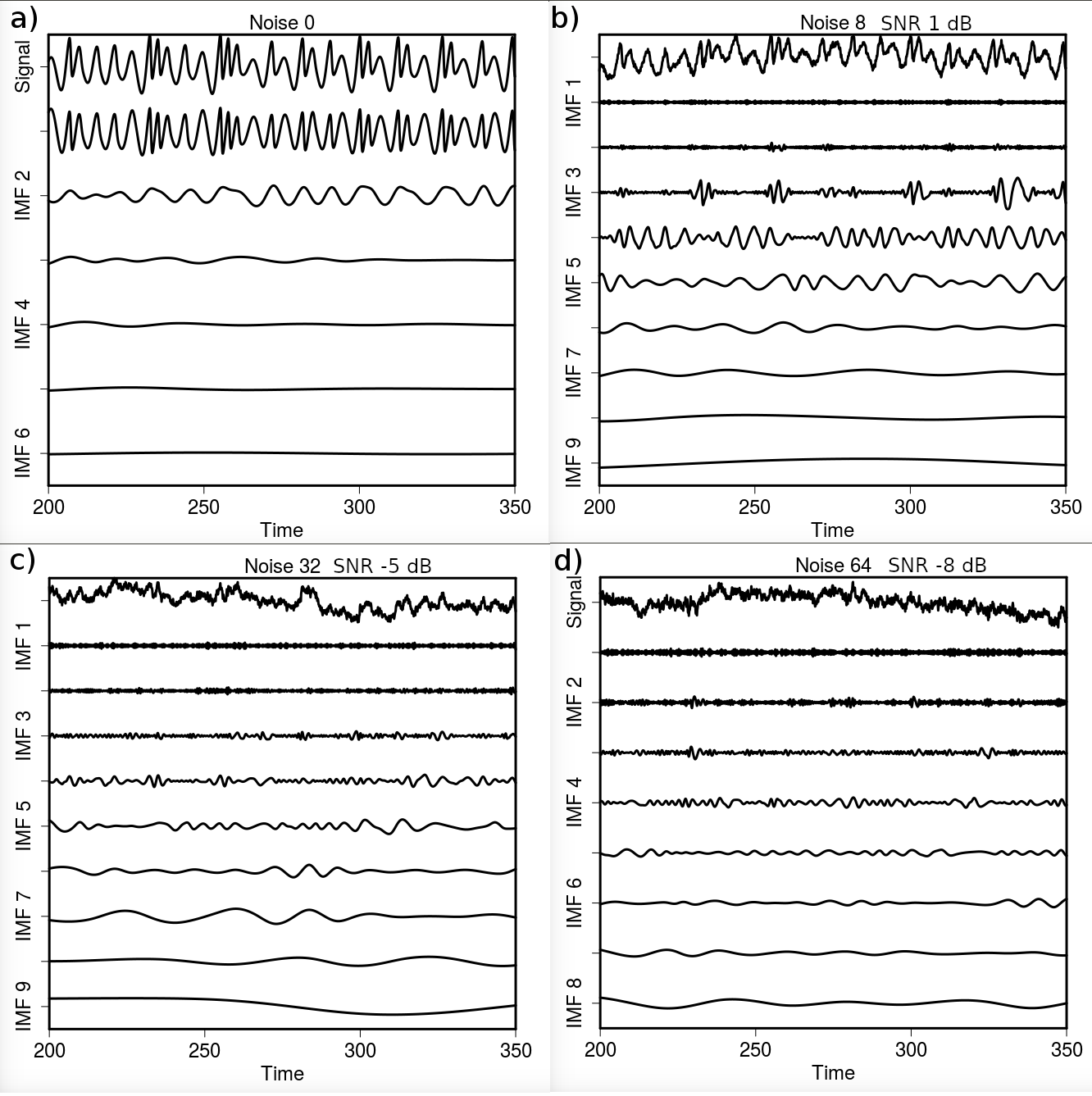}
\caption{Rössler state-space variable $x$ time series with 3 instances of additive noise, and the resultant IMFs.}
\label{fig:RosslerIMF}
\end{figure}

%---------------------------------------------------------
The first step in EMM is to obtain IMFs of the observation time series. Fig. \ref{fig:RosslerIMF} shows the Rössler variable $x$ and it's empirical mode decomposition IMFs at four values of additive multispectral noise amplitude $A$ from equation \ref{equ:Noise}. We can see that IMFs partition timescale variance into modes with decreasing ranges of instantaneous frequency, high frequencies in the low-order modes and lower frequencies in the higher-order modes.  We will leverage this partition as a data-driven filter to select IMFs that better represent the underlying dynamics. 

%---------------------------------------------------------
\subsection{Empirical dynamic modeling}
%---------------------------------------------------------
Empirical dynamic modeling provides state-space forecasting and cross-variable analysis, here, we use the simplex algorithm to project univariate state-space dynamics to assess the fitness of state-space representations derived from traditional time-delay embeddings, and, from IMFs of observed variables.  IMFs are computed over the entire time series, and, partitioned into training library and prediction sets in the same manner as any EDM candidate time series. The forecast variable is $z$ from Eq. \ref{equ:Rosslerz}, with forecasts made from three different state-space representations consisting of vectors $[x,y]$, $[x,y,z]$, or their IMFs.

The simplex algorithm is a state-space projection from a simplex of nearest neighbors in the training data ({\it library}) closest to the state-space of the prediction point.  The simplex consists of $E+1$ points, where $E$ is the embedding dimension in the case the state-space is a time-delay embedding, or, simply the dimension of the multivariate state-space.  Reference \cite{Sugihara1990} describes the simplex algorithm.

%---------------------------------------------------------
\subsubsection{Model comparison}
%---------------------------------------------------------
We examine four different EDM models (Table \ref{table:RosslerEDM}). The first is a full variable state-space of $[x,y,z]$ ($E\!=\!3$) with complete information. This serves as a reference.  Second, a multivariable $[x,y]$ ($E\!=\!2$) state-space representing a naive predictor without information of the forecast variable $z$.  Third, a Takens time-delay state-space of variables $[x,y]$ with embedding dimension $E_e\!=\!3$ resulting in a state-space dimension of $E\!=\!6$, and fourth, multivariate state-space of all IMFs of $[x,y]$.  Here, the dimension $E$ is equal to the number of IMFs.

Since we are interested in the ability of these different representations to model the underlying dynamics, all models use the target time series of noiseless variable $z$, and perform simplex projection at a forecast interval of $T_{p}\!=0$.  All models are evaluated in an out-of-sample prediction with the training library consisting of points [1-2000], and prediction set [2001-3000].

\begin{table}
\centering
\caption{Rössler EDM State-space Variables}
\label{table:RosslerEDM}
\begin{tabular}{lll}
\hline\noalign{\smallskip}
State-space      & Variables        & E \\
\noalign{\smallskip}\hline\noalign{\smallskip}
Multivariable    & $x,y,z$          & 3 \\
Multivariable    & $x,y$            & 2 \\
Takens time-delay& $x,y$            & 6 \\
Multivariable    & $IMF_{x},IMF_{y}$ & $N_{IMF}$ (12-18) \\
\noalign{\smallskip}\hline
\end{tabular}
\end{table}

\begin{figure}
\centering
\includegraphics[width=2.8in]{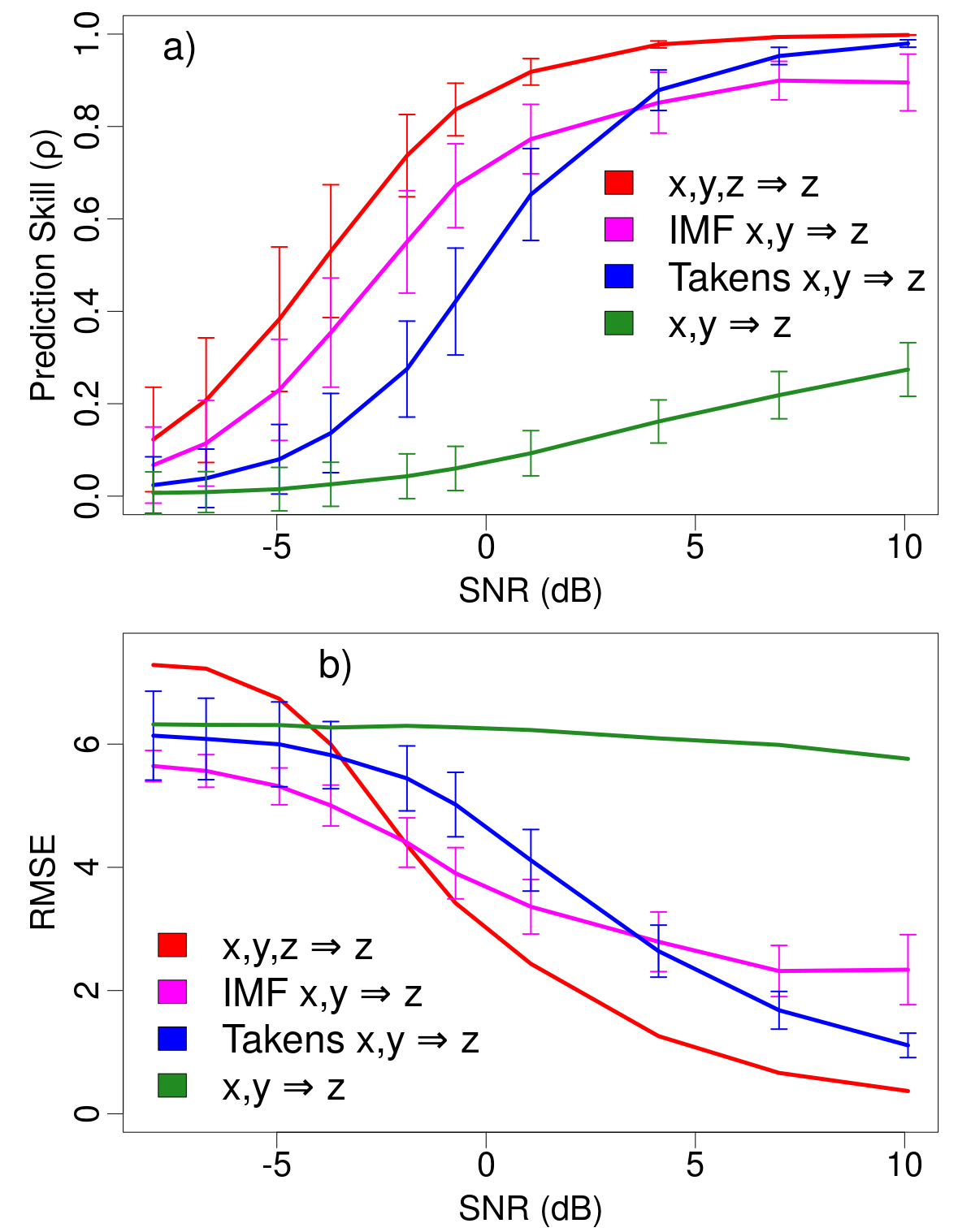}
\caption{Pearson correlation and RMSE of simplex projected Rössler variable $z$ as a function of noise over 1000 noise realisations at each noise amplitude.  Lines are mean values, error bars $\pm$ standard error.}
\label{fig:RosslerEDMNoise}
\end{figure}

Fig. \ref{fig:RosslerEDMNoise} shows ensemble simplex projection results as function of signal-to-noise ratio for the four state-space representations over 1000 noise realisations at each noise amplitude.  Here, we find the naive predictor $x,y\to z$ incapable of meaningful prediction, while the Takens time-delay embedded model (Takens $x,y\to z$) approaches that of the reference model ($x,y,z\to z$) at high signal-to-noise ratios.  At high SNR, above approximately 3 dB, the Takens model outperforms EMM based on IMFs, however, as SNR falls below 3 dB EMM outperforms the time-delay representation providing information recovery closer to the full-information reference model.  A SNR of 3 dB translates to signal power a factor of 2 greater than the noise power.

As noted, at SNR greater than 3 dB, EMM performs slightly worse than the time-delay embedding.  It may be that the lower dimensional time-delay embedding provides a more compact state-space simplex that better represents state evolution when noise is low. Here, the time-delay state-space is 6 dimensional, while the EMM multivariate space is 12 dimensional.

An alternative EDM algorithm: Sequential Locally Weighted Global Linear Maps (S-map) \cite{Sugihara1994} applies a localisation kernel to the library of states, rather than using a fixed number of k-nearest-neighbors to define a simplex. If dimensional inflation of the simplex is the cause of degraded performance at high SNR, S-map based EMM may address this issue.  However, the goal of EMM is to improve state-space models in the presence of noise.

These results suggest that when noise power is more than 1/2 the signal power, even an indiscriminate use of IMFs as EDM state-space variables instead of time-delay embedding can improve cross-variable projection $x,y\to z$, evincing a more inforamtic representation of the underlying dynamics.

%---------------------------------------------------------
\subsubsection{IMF selection}
%---------------------------------------------------------
\label{IMFSelection}
It was observed that low order IMFs capture high frequency noise not present in the underlying Rössler dynamics, and, that use of all IMFs as EDM state-space variables improves predictability in the presence of noise.  It is then natural to ask whether subsets of IMFs can provide additional gains in state-space representation and model fidelity, or, provide equitable predictability from a lower-dimensional representation.

One method to identify noise-dominated IMFs is to examine the variance of IMF instantaneous frequencies (IF) since high variance in instantaneous frequency is indicative of noise, while low variance suggests a relatively stable oscillatory mode reflective of low-dimensional dynamics.  The idea being to discard IMFs with instantaneous frequency variance above some threshold.  However, this does not set a useful bound on low frequency IMFs, and requires a data-specific threshold selection. 

Following the model agnostic, data driven approach wherein model predictability serves as a metric of information content, Ye \& Sugihara suggested multiview embedding as an EDM algorithm to select maximally informative state-space variables \cite{Ye2016}.  Multiview embedding evaluates model fidelity across all combinations of a set of candidate variables, selecting the $D$-dimensional subset with the highest predictive skill. In standard multiview embedding, the selection process is performed with in-sample simplex projection, the top ranked variables used in an out-of-sample prediction for the final model output.  However, in-sample ranking poses a problem with IMFs since low-frequency IMFs approach, and eventually express monotonic functions.

Simplex projection is a nonlinear state-space mapping from state-space variables to a target without constraint on the variables. If the variables support a unique mapping across their domain, then good {\it in-sample} predictability can be achieved. This means that arbitrary, non-constant, non-oscillatory functions that provide a unique multivariable mapping can be mapped in-sample with good fidelity to the target.  Since low frequency IMFs have very little or no oscillatory content, they are likely to be selected as in-sample variables with high predictive skill.  Normally, candidate state-space variables are oscillatory or physically relevant, and this is not a concern.  Here, we use all available IMFs, and address IMF selection by modifying standard multiview embedding to use {\it out-of-sample} predictability as the metric for ranking top predictors.

The modified multiview embedding is performed using the complete set of IMFs from variables $x$ and $y$ with the training library of points [1-2000], prediction set [2001-3000], and forecast interval $T_P = 0$. The embedding dimension is set to $E = 1$ so that the IMFs are not time-delay embedded, but used explicitly as state-space variables. The most predictive combination of $D$ IMFs are then selected to create a multivariable state-space.

\begin{figure}
\centering
\includegraphics[width=3in]{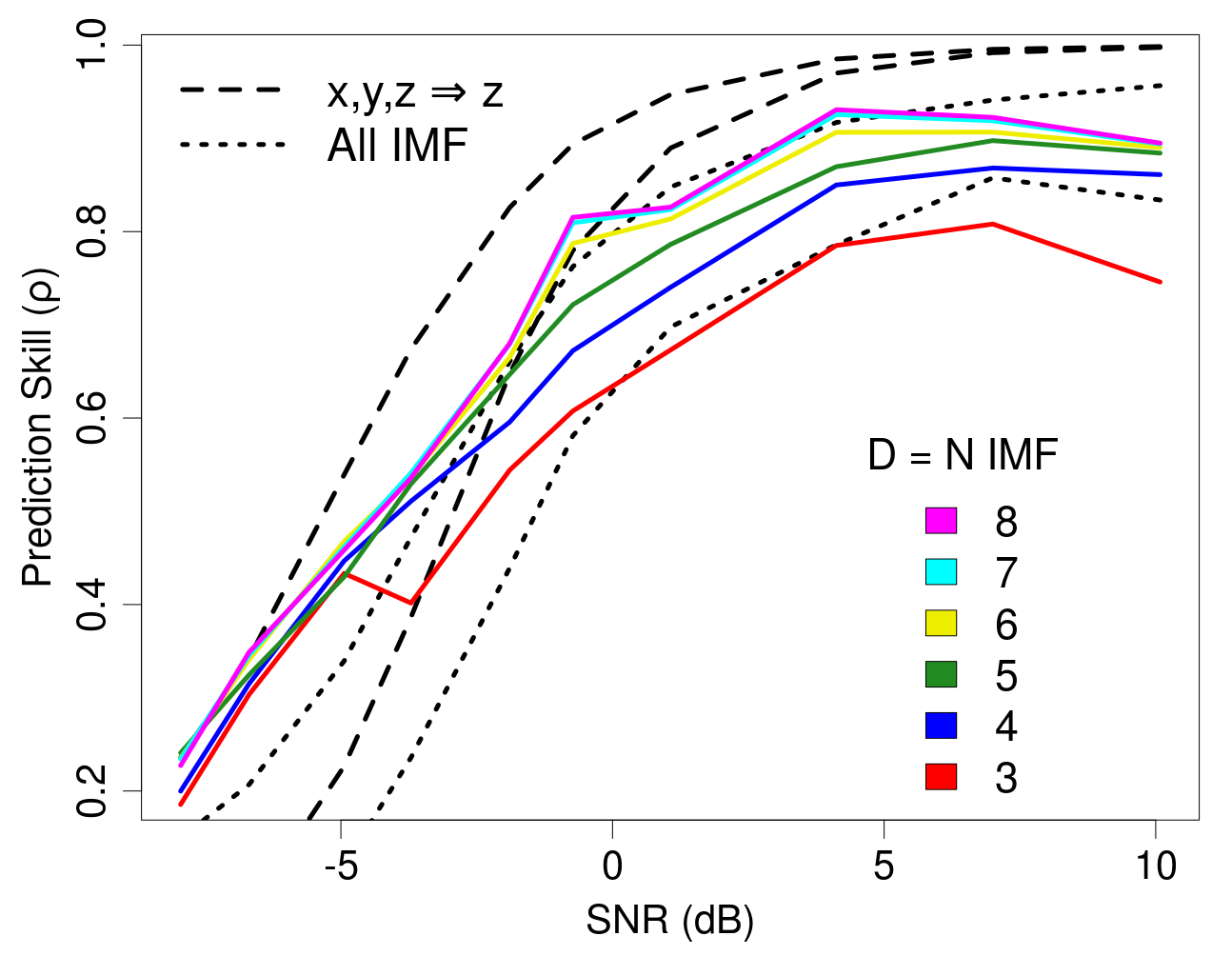}
\caption{Simplex projection skill of the most predictive IMFs of one noise realisation as a function of $D$, the number of IMFs. Dashed lines are the $\pm$ standard error envelopes of the 1000 noise ensembles of the reference model and state-spaces using all IMFs.}
\label{fig:MultiviewNoiseIMF}
\end{figure}

Fig. \ref{fig:MultiviewNoiseIMF} plots simplex projection skill of the most predictive IMFs determined by multiview embedding for state-space dimensions of $D$ from 3 to 8, comparing them to the $\pm$ standard error envelopes of the 1000 noise ensembles of the reference model and state-spaces using all IMFs.  At SNR immediately below 3 dB the predictive skill of EMM with state-space dimension of $D = 6$ has converged to match the upper envelope of the full IMF state-space ensembles. At SNR below 0 (noise power exceeds signal power) the $D = 6$ projection exceeds that of the full IMF state-space.

To summarise, an $E = 3$ dimensional time-delay embedding on variables $x$ and $y$ to simplex project $z$ from a state-space with dimension $D$ = 6 matches that of the fully informed reference model ($[x,y,z]$ projecting $z$) at zero noise.  As noise increases the Takens state-space model degrades in predictive skill, and, both an indiscriminate use of all IMFs from noisy $x$ and $y$, as well as a selected subset of $D = 6$ IMFs outperforms the Takens model at SNR less than 3 dB.  We infer that EMM can provide improved state-space representations in the presence of noise.

%---------------------------------------------------------------------------
\section{Salinity in Florida Bay}
\begin{figure}
\centering
\includegraphics[width=2.8in]{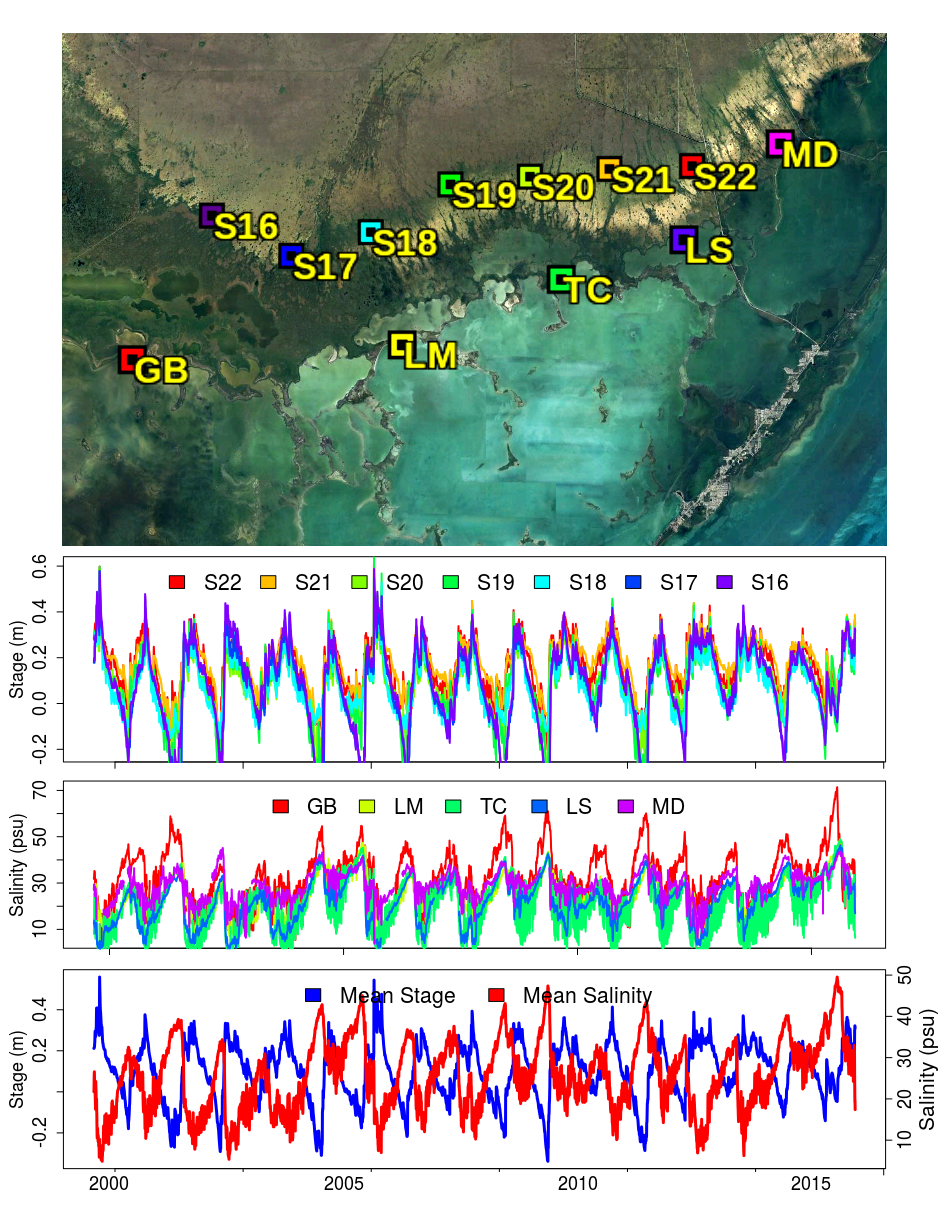}
\caption{Top: Aerial photograph of the southern Everglades and Florida Bay with hydrographic monitoring stations. Bottom: Station water levels (stage), salinity, and station-averaged water level and salinity over 16 years.}
\label{fig:FLBay}
\end{figure}

%---------------------------------------------------------------------------
Coastal South Florida is fringed by national parks including Biscayne and Everglades National Parks.  Florida Bay is situated between the southern edge of the Florida peninsula and Florida Keys, is part of Everglades National Park, and, is an ecologically important and proliferent multispecies marine and estuarine nursery. Hydrologically, Florida Bay is interesting in that annual evaporation exceeds rainfall, resulting in widely varying coastal salinities as shown in Fig. \ref{fig:FLBay}.  Hypersalinity is a seasonal occurrence, with episodic extremes surpassing 50 psu.  These extreme hypersalinity events are associated with ecological collapse \cite{Hall2016, Johnson2018}. Clarification of influence variables and the ability to forecast salinities will directly inform ecosystem resource management.

Our objective is to forecast salinity in Garfield Bight (GB), assessing predictive performance of EDM simplex using three different state-space representations described below.  The Florida Bay data consist of 6332 days of mean water level (stage), salinity, evaporation, and rainfall.  The lower panel of Fig. \ref{fig:FLBay} illustrates a nonlinear inverse relationship between salinities and coastal marsh water levels signifying that water levels in the southern Everglades are a prime determinant of coastal salinities.  Other causal variables are known to include evaporation, which is related to temperature and solar radiation, rainfall, water levels in the bay, and, previous salinity.

An initial step in EDM analysis is to examine the dimensionality and nonlinearity of the data \cite{Chang2017}, here, indicating that the data are nonlinear, and, a Takens time-delay embedding dimension of 5 provides good model predictability.  Results of this data discovery are presented in appendix 1 (section \ref{appendix:FLBay}).

%---------------------------------------------------------------------------
\subsection{State-space models}
%---------------------------------------------------------------------------
\label{stateSpaceModels}
The first state-space model is a Takens $E\!=\!5$ time-delay embedding of GB salinity.  Second, a multivariate state-space is constructed from the 5 variables GB salinity, stage, evaporation, rainfall, and, coastal land water level at station S16.  These variables are known influencers of salinity as determined by the Florida Bay Assessment model \cite{Park2016}.  The third state-space consists of selected IMFs of GB salinity, water level, evaporation, and coastal land water level at S16.  IMFs are shown in appendix 1 (section \ref{appendix:FLBay}).

%---------------------------------------------------------------------------
\subsubsection{Model comparison}
%---------------------------------------------------------------------------
The comparative state-space analysis defines an in-sample data set as days [1-6330] for the training library, and days [5967-6275] as the prediction set. This prediction set corresponds to the period 2016-01-01 to 2016-11-04. The EMM model uses IMFs 5,6,7 of the salinity and stage variables, and IMFs 6,7,8 for S16 water level and evaporation (figure \ref{fig:GB_IMF}).  These IMFs ignore high frequency, noise dominated components, as well interannual variations not deemed important on daily timescales.  

\begin{figure}
\centering
\includegraphics[width=3.3in]{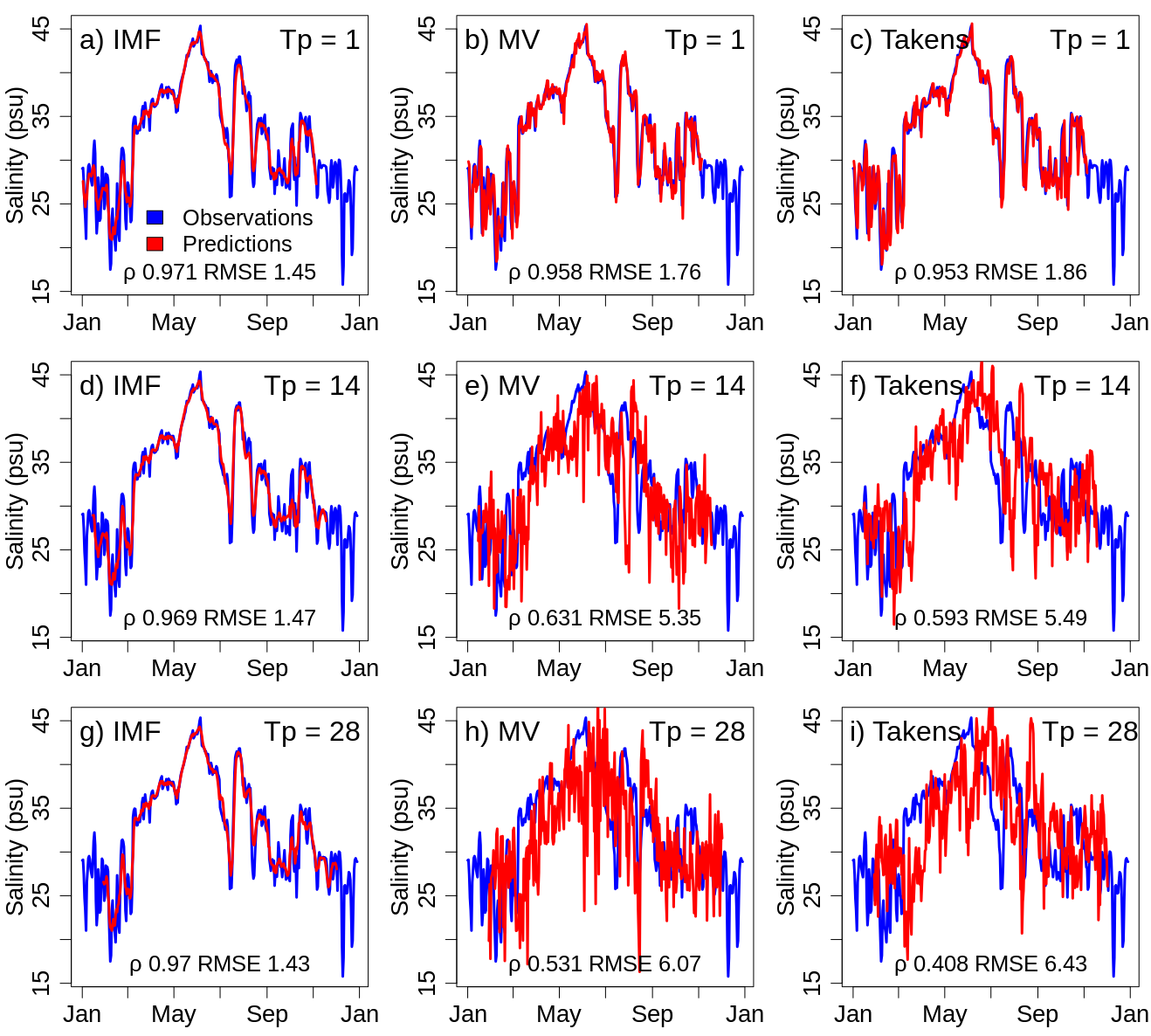}
\caption{Simplex in-sample projections of GB salinity for the year 2016 at different prediction intervals $T_p$ for the EMM, multivariable, and Takens time-delay state-space representations.}
\label{fig:Predict_IMF_MV_Tp}
\end{figure}

Fig. \ref{fig:Predict_IMF_MV_Tp} plots a series of in-sample GB salinity projections for the three models over a sequence of prediction horizons $T_p$.  It is worth reinforcing that these are not predictions \textit{per-se}, since the model is purely in-sample, rather, we seek to evaluate internal model representations of the dynamics.  Here we see that while the $T_p\!=\!1$ projections are good for all three models, the EMM model misses high frequency variations evident in the January to March and September to October time frames.  As prediction interval increases, these high frequency contributions seem to dominate the multivariate and time-delay embedding projections, significantly degrading accuracy, while the EMM model maintains a high fidelity internal representation of the dynamics. 

%---------------------------------------------------------------------------
\subsection{Forecasting}
\begin{figure}
\centering
\includegraphics[width=2.5in]{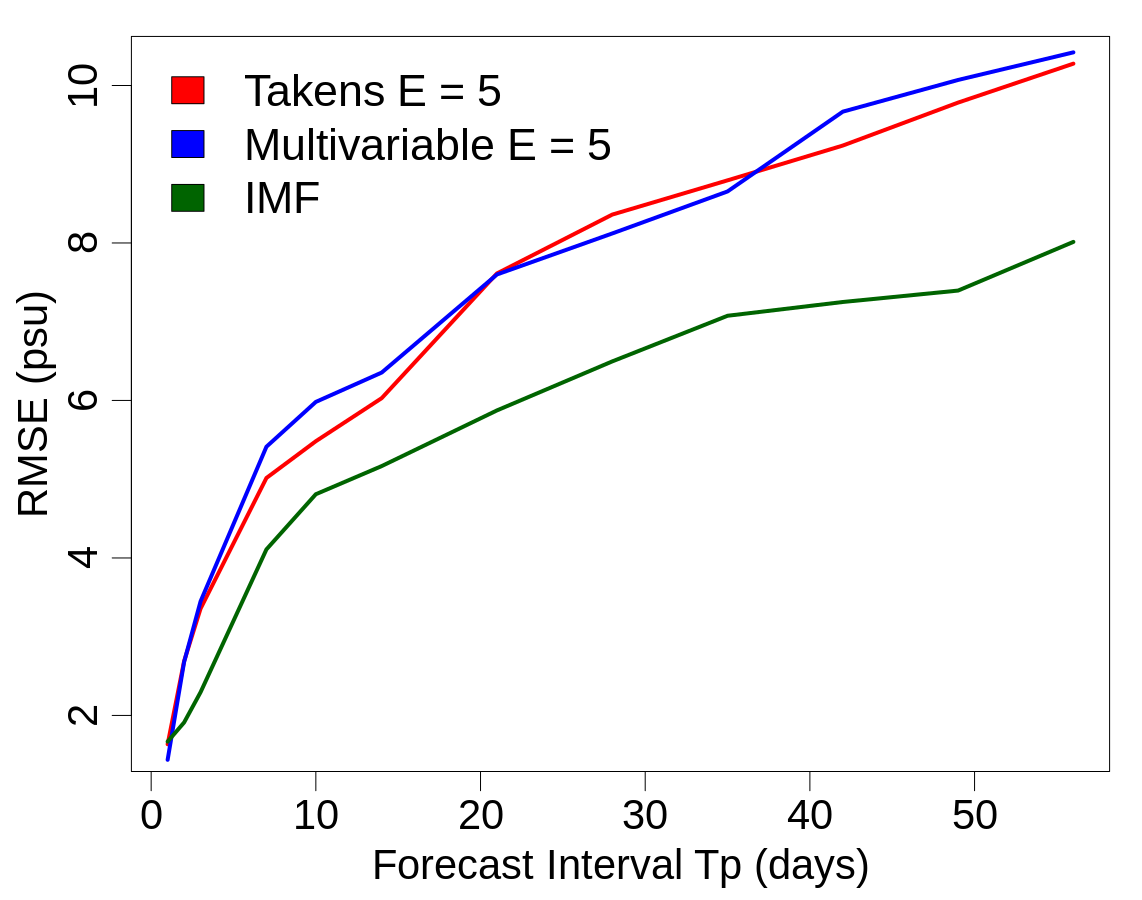}
\caption{Root mean square error (RMSE) of GB salinity out-of-sample simplex forecasts for the year 2016 at different prediction intervals $T_p$ for the EMM, multivariable, and time-delay state-space representations.}
\label{fig:Stat_MovingPred}
\end{figure}

%---------------------------------------------------------------------------
We now turn to the question of whether EMM can be useful in an out-of-sample regime which is requisite for an applied forecast system. Simplex forecasts are made with the three state-space models from a training library consisting of points [1-5475] expanded in 30 day increments starting at day 5475 (2014-08-27) until the end of the record. Predictions are made at forecast intervals of 1 to 56 days in 7 day increments, starting at one day past the end of the training library.  This 30-day moving window provides 28 samples at each prediction interval.  The mean RMSE over each prediction interval is shown in Fig. \ref{fig:Stat_MovingPred} indicating that the EMM IMF state-space provides generally more accurate predictions of salinity than the multivariate and time-delay state-space models.

%---------------------------------------------------------------------------
\subsubsection{Comparison to numerical model}
%---------------------------------------------------------------------------
As noted earlier, salinity in coastal Florida Bay is an ecologically important factor, linked to the onset of widespread ecological collapse \cite{Hall2016, Johnson2018}.  In 2015, environmental conditions aligned to produce historic record maximum salinities in excess of 70 psu (seawater is nominally 35 psu) followed by widespread sea-grass mortality and food-web disruption.  The ability to forecast such events, or to attribute causal variables, provides actionable information for natural resource managers.  Accordingly, equation-based physical models have been developed to explore these issues, and, the Bay Assessment Model (BAM) is one of the latest and best performing salinity models \cite{Park2016}. Here, we compare BAM and EMM forecasts for the 2015 hypersalinity event.

EMM forecasts are out-of-sample based on progressive training library and prediction horizon times.  The training library starts with days [1-5721]. Predictions are made for 121 days starting at day 5722 (2015-5-1) over a range of prediction horizons $T_p$ from 1 to 21 days. After each set of predictions, the training library end day and prediction start day are incremented by 1 day.

\begin{figure}[H]
\centering
\includegraphics[width=3in]{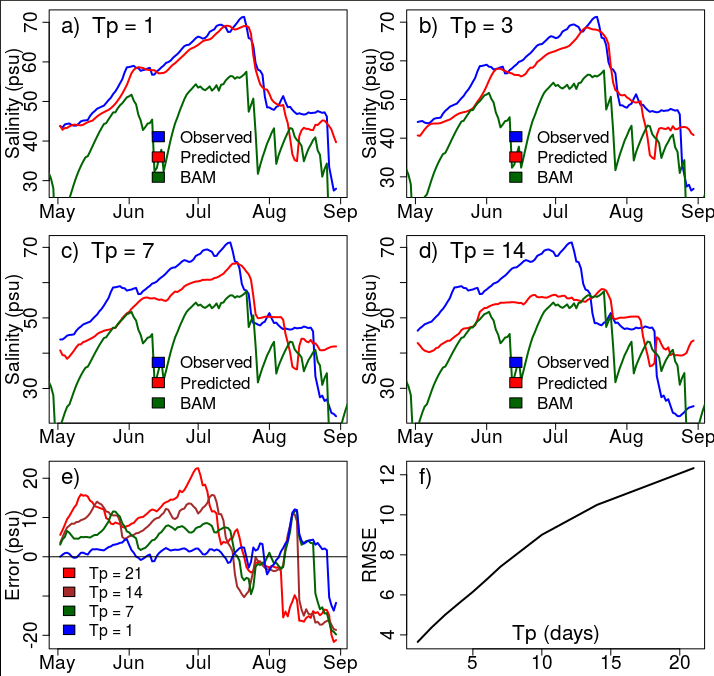}
\caption{Salinity forecasts in Garfield Bight during the 2015 hypersalinity event. \:a) - d) EMM forecasts (red) with observed data (blue) and results from the Bay Assessment Model (green). \:e) EMM forecast error for different prediction horizons $T_p$. \:f) EMM RMS error over the prediction period May 1 - August 31, 2015 as a function of forecast interval $T_p$.}
\label{fig:GB_Forecast_2015}
\end{figure}

Fig. \ref{fig:GB_Forecast_2015} presents comparisons of the EMM forecasts with BAM results, where we note that BAM fails to capture the extreme hypersalinity event of 2015.  EMM forecasts provide accurate salinity predictions at $T_p\!=\!1$ with good forecasting of extreme salinity.  Forecasts remain reasonably good at $T_p\!=\!3$, having lost good fidelity at $T_p\!=\!7$, but still outperforming the numerical model.  This suggests that the method of progressive EMM projections at $T_p\!=\!1$ are good candidates for an operational forecast system. 

%---------------------------------------------------------------------------
\section{Conclusion}
%---------------------------------------------------------------------------
The 21$\mathrm{^{st}}$ Century revolution in data collection and analysis has fundamentally changed our ability to infer relationships and dynamical behavior of complex systems. A powerful corollary is the emergence of data-driven, model-free analytical techniques where presumptions and models are not fit to observations, but where the data itself is represented in a state-space facilitating the discovery of dynamical relationships enabling state forecasting.

Inevitably, observational data are intertwined with noise, with potential to disrupt accurate state-space representations. Such disruptions motivate us to suggest empirical mode modeling (EMM), a synthesis of empirical mode decomposition (EMD) and empirical dynamic modeling (EDM).  As the names suggest, these are data-driven, model-free techniques ideally suited for nonlinear dynamical analysis. EMM uses EMD intrinsic mode functions (IMF) as state-space vectors in the EDM framework. Since IMFs naturally decompose time series into physically-relevant modes, they are well-suited for isolation and removal of noise.

EMM was assessed by application to a synthetic data set with controlled SNR, and, it was found that for high SNR state-spaces EMM as implemented here does not improve projection skill.  However, as SNR falls below 3 dB EMM is found to improve model predictability and information content of the state-space, leading us to anticipate that noise-dominated signals can benefit from EMM. 

Application to a geophysical data set composed of inherently noisy data found that EMM provided superior forecasting ability in comparison to time-delay or multivariate state-space representations.  Further, the EMM representation outperformed the best available numerical model in predicting the 2015 hypersalinity event in Garfield Bight of Florida Bay, an extreme occurrence without precedent in the observational record. 

Collectively, this work indicates that EMM can be a useful technique to improve state-space representations in the presence of noise with signal-to-noise ratios less than 3 dB. 

Although this initial implementation and assessment of EMM seems promising, there are inherent limitations in the method, and many avenues for improvement.  The primary limitation is that EMD is not applicable to non-oscillatory signals.  Time series such as neuronal spike trains, rainfall, or other discrete, Poisson process dynamics will not be amenable to EMM based on EMD IMFs.  This might be addressed by replacing the EMD IMF with a modal decomposition that captures non-oscillatory time-dependence, for example, a discrete wavelet transform.

Regarding improvements, investigating methods to optimally select IMFs to avoid noise, or, to maximize cross-variable information assessments are warranted.  Here, we have relied on EDM multiview embedding.  Second, EMD has been extended beyond the original decomposition algorithm used here.  For example, ensemble empirical mode decomposition (EEMD) has been found to mitigate mode-mixing and improve physical significance of modes \cite{Wu2009}.  EMD has also been extended into hybrid machine learning paradigms \cite{Chen2019, Jiang2019} and multivariate implementations \cite{Rehman2009}.

Another obvious exploration is to assess complementary mixed embeddings as a state-space representation.  For example, combinations of Takens time-delay embedded vectors with IMFs, and, with multivariate observations.  Yet another is to use other nonlinear modal decompositions such as the discrete wavelet transform either in-place of, or, as a complement to IMFs.  Finally, the use of sequential locally weighted global linear maps (S-maps)\cite{Sugihara1994} may provide improved state-space representation and forecast skill.

%\begin{acknowledgements}
%\end{acknowledgements}

\section*{Conflict of interest}
The authors declare that they have no conflict of interest.

\clearpage
%---------------------------------------------------------------------------
% Appendixes appear before the acknowledgment.
%---------------------------------------------------------------------------
%\vspace{-1.2cm} % ? A . is printed if \appendices used. Back up to cover it.
%---------------------------------------------------------------------------
\section{Appendix 1: Florida Bay salinity data discovery}
%---------------------------------------------------------------------------
\label{appendix:FLBay}
Application of EDM to a dataset is predicated on assessing whether the underlying dynamics are state-dependent (nonlinear), and if so, estimating an embedding dimension to use if univariate data are to be time-delay embedded.  A primer on EDM analysis can be found in \cite{Chang2017}.

The Florida Bay data consist of 6332 daily values (September 1 1999 though December 31 2016). We use days 1-5000 as the training (library) set, and days 5001-6300 as the prediction. 

\begin{figure}[H]
\centering
\includegraphics[width=2.5in]{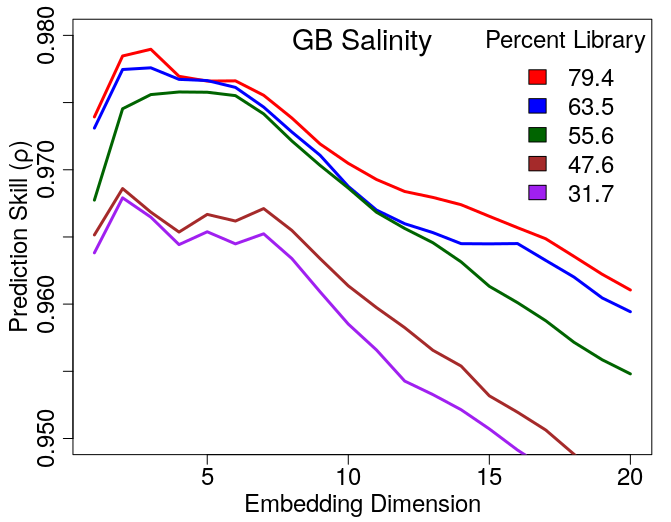}
\caption{Out-of-sample simplex model correlation for Florida Bay station GB salinity as a function of time-delay embedding dimension.  Different curves are based on different library sizes, expressed as a percentage of the total data size. The maximum library index is 5000, predictions are over days 5001-6300.}
\label{fig:GB_salt_E}
\end{figure}

Fig. \ref{fig:GB_salt_E} shows simplex prediction skill as a function of Takens embedding dimension for Garfield Bight (GB) salinity.  Predictions are made at a forecast interval of $T_p\!=\!1$ timestep ahead.  This suggests that embedding dimensions in the range of 2-6 provide the best model fidelity, and that the underlying dynamics of the salinity are low-dimensional. 

\begin{figure}%[H]
\centering
\includegraphics[width=2.5in]{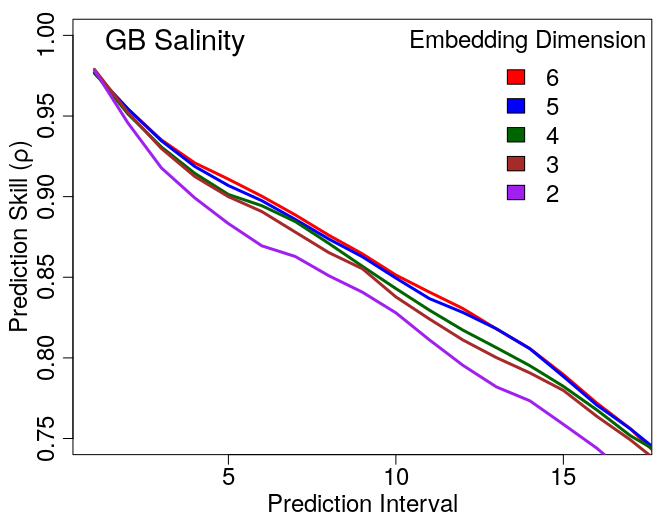}
\caption{Out-of-sample simplex model correlation for Florida Bay station GB salinity as a function of forecast interval of $T_p$.}
\label{fig:GB_salt_Tp}
\end{figure}

Fig. \ref{fig:GB_salt_Tp} presents simplex prediction skill as a function of forecast interval $T_p$ at different embedding dimensions $E$.  The saturation of predictability as $E$ approaches 5 indicates that a dimension of 5 can be a reasonable choice for EDM analysis of this data.  The decline in predictability as $T_p$ increases is consistent with dynamics of a nonlinear system. 

\begin{figure}%[H]
\centering
\includegraphics[width=2.5in]{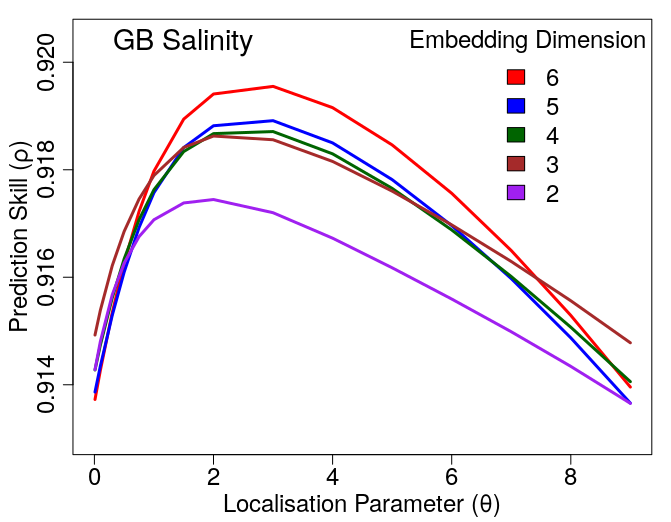}
\caption{Out-of-sample s-map model correlation for Florida Bay station GB salinity as a function of nonlinear localization parameter $\theta$.}
\label{fig:GB_salt_Theta}
\end{figure}

Interpreting nonlinearity as state-dependence \cite{Sugihara1994}, the EDM s-map algorithm allows one to assess nonlinearity with examination of predictive skill as a function of state-space linear localization ($\theta$).  Fig. \ref{fig:GB_salt_Theta} shows this assessment for the GB salinity data, indicating a weak, but robust state-dependence, again indicative that the salinity dynamics are nonlinear.

\begin{figure}%[H]
\centering
\includegraphics[width=0.9\textwidth]{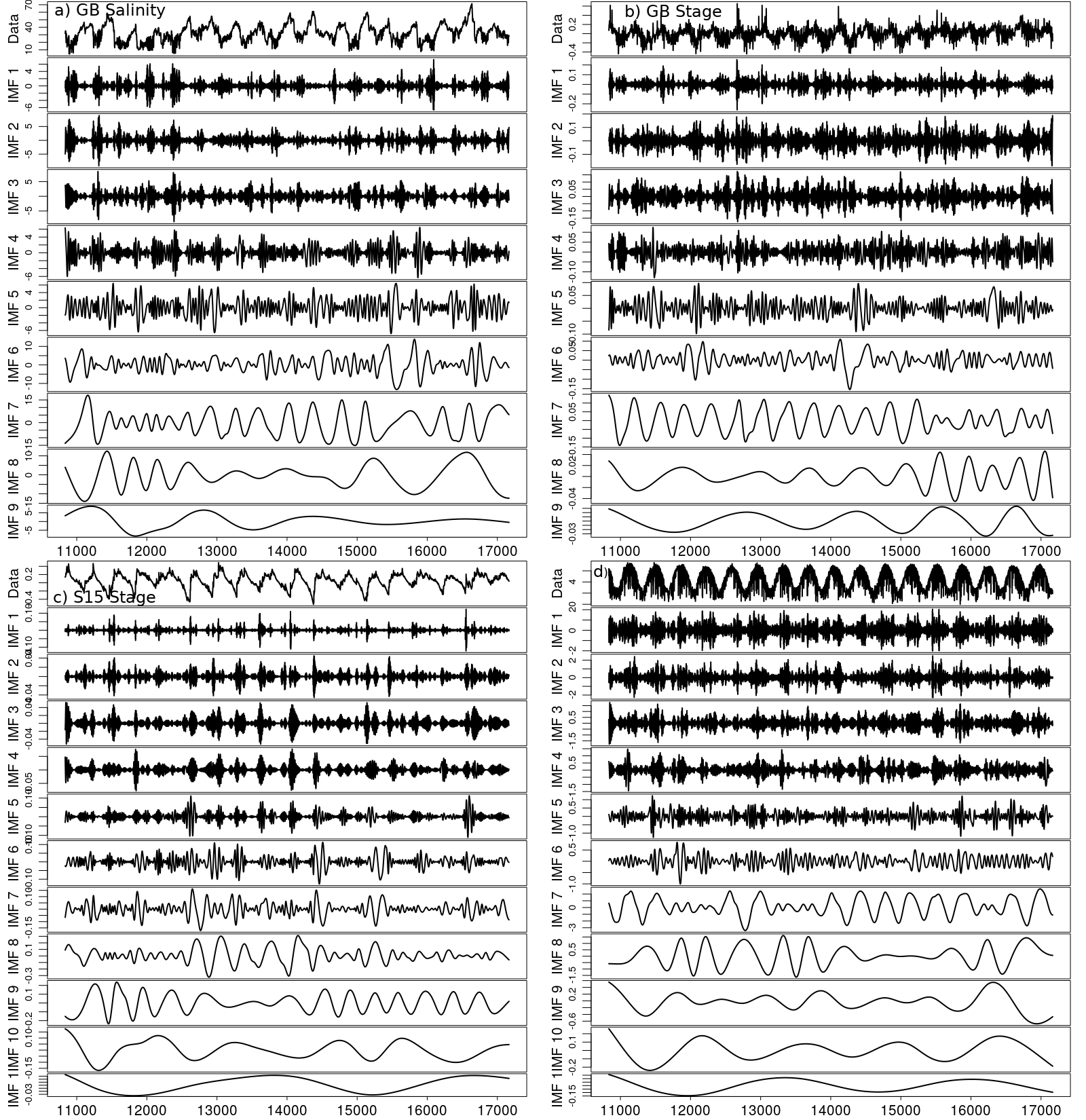}
\caption{EMD IMFs of variables influencing Garfield Bight salinity. \:a) Salinity. \:b) Water level. \:c) Coastal land water level. \:d) Evaporation.}
\label{fig:GB_IMF}
\end{figure}

Fig. \ref{fig:GB_IMF} presents EMD IMFs of physical variables influencing salinity in Garfield Bight, Florida Bay.  Low order, high frequency IMFs are considered noise dominated and not used as EDM state-space vectors. High order, low frequency IMFs are also not used as state-space variables. 

\clearpage
%---------------------------------------------------------------------------

\end{document}